\documentclass{article} 
\usepackage{iclr2021_conference,times}

\usepackage{amsmath,amsfonts,bm}

\def\eqref#1{equation~\ref{#1}}

\def\1{\bm{1}}

\DeclareMathAlphabet{\mathsfit}{\encodingdefault}{\sfdefault}{m}{sl}
\SetMathAlphabet{\mathsfit}{bold}{\encodingdefault}{\sfdefault}{bx}{n}

\DeclareMathOperator*{\E}{\mathbb{E}}

\newcommand{\softmax}{\mathrm{softmax}}

\usepackage{hyperref}
\usepackage{url}
\usepackage{amsmath,amssymb} 
\usepackage{mathtools}
\usepackage{bbm}
\usepackage{xfrac}
\usepackage{color}
\usepackage{caption}
\usepackage{sidecap}
\usepackage{graphicx}
\usepackage{subfigure}
\usepackage{makecell}

\usepackage{adjustbox}
\usepackage{array}

\newcolumntype{R}[2]{%
    >{\adjustbox{angle=#1,lap=\width-(#2)}\bgroup}%
    l%
    <{\egroup}%
}

\newcommand{\minisection}[1]{\textbf{#1}\hspace{0.3em}}

\newcommand{\loss}{L}
\newcommand{\nll}{\ell}

\newcommand{\normal}{\mathcal{N}}
\newcommand{\lp}{\phi}  

\DeclarePairedDelimiter{\abs}{\lvert}{\rvert}

\newcommand{\power}{\alpha}

\newcommand{\lossfun}[1]{ \rho\left( #1 \right)}

\newcommand{\prob}[1]{p\left( #1 \right)}
\newcommand{\partition}[1]{Z\left( #1 \right)}

\title{It Is Likely That Your Loss \\ Should be a Likelihood}

\author{Mark Hamilton \\
MIT, Microsoft\\
\texttt{markth@mit.edu} \\
\And
Evan Shelhamer \\
MIT, Adobe Research \\
\texttt{shelhamer@adobe.com} \\
\And
William Freeman \\
MIT, Google \\
\texttt{billf@mit.edu}
}

\iclrfinalcopy 
\begin{document}

\maketitle

\begin{abstract}
Many common loss functions such as mean-squared-error, cross-entropy, and reconstruction loss are unnecessarily rigid. Under a probabilistic interpretation, these common losses correspond to distributions with fixed shapes and scales. We instead argue for optimizing full likelihoods that include parameters like the normal variance and softmax temperature.
Joint optimization of these ``likelihood parameters'' with model parameters can adaptively tune the scales and shapes of losses in addition to the strength of regularization.
We explore and systematically evaluate how to parameterize and apply likelihood parameters for robust modeling, outlier-detection, and re-calibration.
Additionally, we propose adaptively tuning $L_2$ and $L_1$ weights by fitting the scale parameters of normal and Laplace priors and introduce more flexible element-wise regularizers.
\end{abstract}

\section{Introduction}
\label{sec:intro}

Choosing the right loss matters.
Many common losses arise from likelihoods, such as the squared error loss from the normal distribution , absolute error from the Laplace distribution, and the cross entropy loss from the softmax distribution.
The same is true of regularizers, where $L_2$ arises from a normal prior and $L_1$ from a Laplace prior.

Deriving losses from likelihoods recasts the problem as a choice of distribution which allows data-dependant adaptation. 
Standard losses and regularizers implicitly fix key distribution parameters, limiting flexibility.
For instance, the squared error corresponds to fixing the normal variance at a constant.
The full normal likelihood retains its scale parameter and allows optimization over a parametrized set of distributions.
This work examines how to jointly optimize distribution and model parameters to select losses and regularizers that encourage generalization, calibration, and robustness to outliers.
We explore three key likelihoods the normal, softmax, and the robust regressor likelihood $\rho$ of \cite{barron2019loss}. Additionally, we cast adaptive \emph{priors} in the same light and introduce adaptive regularizers. In summary:

\begin{enumerate}
 \itemsep-.2em 
  \item We systematically survey and evaluate global, data, and predicted likelihood parameters and introduce a new self-tuning variant of the robust adaptive loss $\rho$
  \item We apply likelihood parameters to create new classes of robust models, outlier detectors, and re-calibrators. 
  \item We propose adaptive versions of $L1$ and $L2$ regularization using parametrized normal and Laplace priors on model parameters.
\end{enumerate}

\section{Background}
\label{sec:background}

\minisection{Notation}
We consider a dataset $\mathcal{D}$ of points $x_i$ and targets $y_i$ indexed by $i \in \{1, \dots, N\}$. 
Targets for regression are real numbers and targets for classification are one-hot vectors. 
The model $f$ with parameters $\theta$ makes predictions $\hat{y_i} = f_\theta(x)$.
A loss $L(\hat{y}, y)$ measures the quality of the prediction given the target.
To learn model parameters we solve the following loss optimization:
\begin{equation}
\min_\theta \E_{(x, y) \sim \mathcal{D}} \loss(\hat{y} = f_\theta(x), y)
\end{equation}A likelihood $\mathcal{L}(\hat{y} | y, \lp)$ measures the quality of the prediction as a distribution over $\hat{y}$ given the target $y$ and likelihood parameters $\lp$.
We use to the negative log-likelihood $\nll$ (NLL), and the likelihood interchangeably since both have the same optima.
We define the full likelihood optimization:
\begin{equation}
\min_{\theta,\lp} \E_{(x, y) \sim \mathcal{D}} \nll(\hat{y} = f_\theta(x) | y, \lp)
\end{equation}
to jointly learn model and likelihood parameters. ``Full'' indicates the inclusion of $\lp$, which control the distribution and induced NLL loss.
We focus on full likelihood optimization in this work.
We note that the target, $y$, is the only supervision needed to optimize model and likelihood parameters, $\theta$ and $\lp$ respectively. Additionally, though the shape and scale varies with $\lp$, reducing the error $\hat{y} - y$ always reduces the NLL for our distributions.

\begin{figure}[t]
\centering     
\subfigure[The Normal PDF and NLL]{
    \includegraphics[width=.48\linewidth]{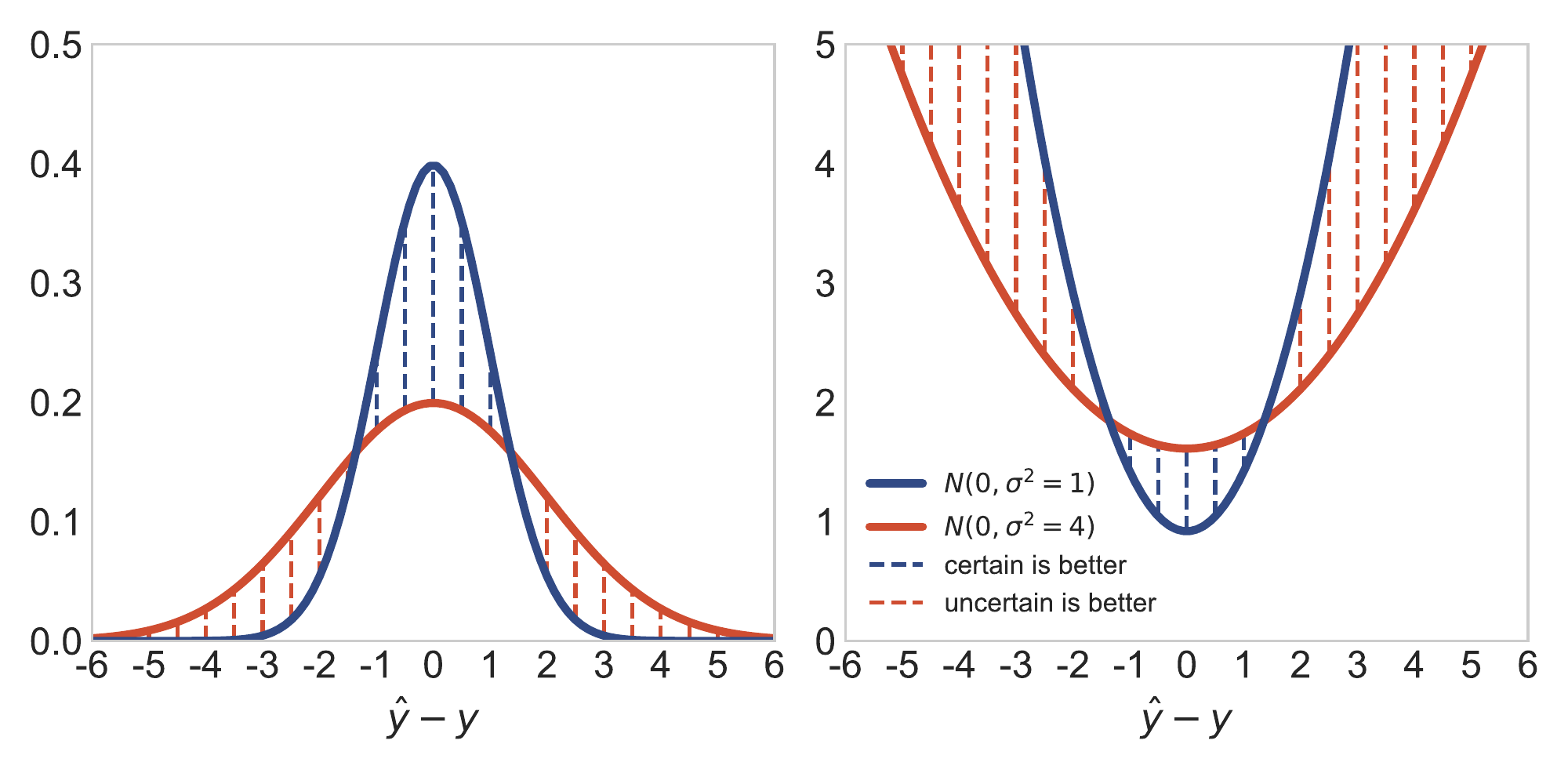}
}
\subfigure[The Softmax CDF and NLL]{
    \includegraphics[width=.47\linewidth]{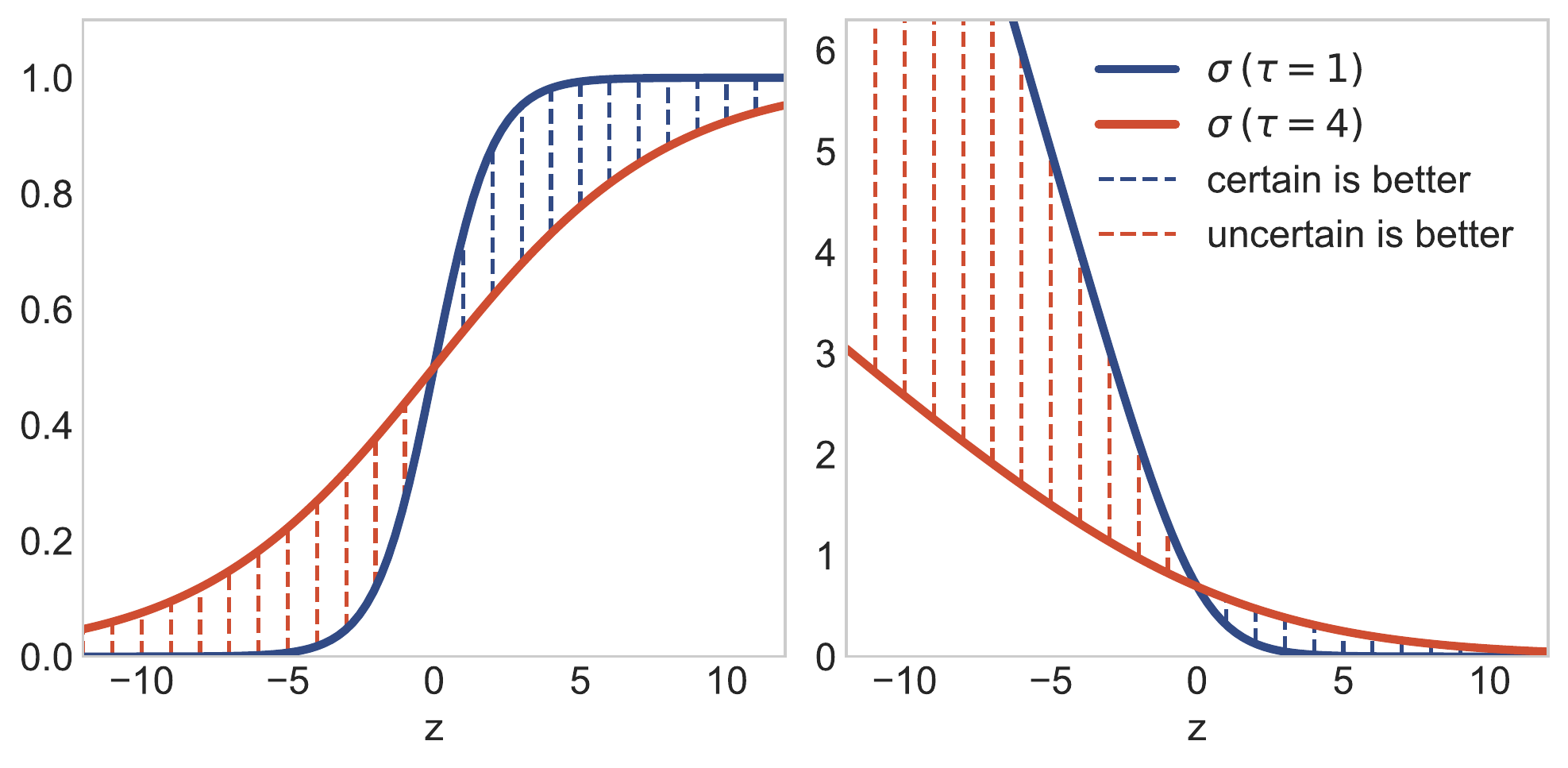}
}
\caption{
    Optimizing likelihood parameters adapts the loss without manual hyperparameter tuning to balance accuracy and certainty.
}
\label{fig:loss-likelihood}
\end{figure}

\minisection{Distributions Under Investigation}
This work considers the normal likelihood with variance $\sigma$ \citep{bishop2006pattern,hastie2009elements}, the softmax likelihood with temperature $\tau$ \citep{hinton2015distilling}, and the robust likelihood $\rho$ \citep{barron2019loss} with shape $\alpha$ and scale $\sigma$ that control the scale and shape of the likelihood.
We note that changing the scale and shape of the likelihood distribution is not ``cheating'' as there is a trade-off between uncertainty and credit. Figure \ref{fig:loss-likelihood} shows how this trade-off affects the Normal and softmax distributions and their NLLs.

The normal likelihood has terms for the residual $\hat{y} - y$ and the variance $\sigma$ as
\begin{equation}
\label{eq:like-normal}
\normal(\hat{y} | y, \sigma) = (2\pi\sigma^2)^{-\frac{1}{2}} \exp\left(-\frac{1}{2}\frac{(\hat{y} - y)^2}{\sigma^2}\right), 
\end{equation}
with $\sigma \in (0, \infty)$ scaling the distribution. The normal NLL can be written 
$\nll_{\normal} = \frac{1}{2\sigma^2}(\hat{y} - y)^2 + \log\sigma$,
after simplifying and omitting constants that do not affect minimization.
We recover the squared error  by substituting $\sigma = 1$.

The softmax defines a categorical distribution defined by scores $z$ for each class $c$ as
\begin{equation}
\label{eq:like-softmax}
\softmax(\hat{y} = y | z, \tau) = \frac{e^{z_y\tau}}{\sum_{c}e^{z_{c}\tau}}, 
\end{equation}
with the temperature, $\tau \in (0, \infty)$, adjusting the entropy of the distribution. The softmax NLL is
We recover the classification cross-entropy loss, $-\log p(\hat{y} = y)$, by substituting $\tau = 1$ in the respective NLL.
We state the gradients of these likelihoods with respect to their $\sigma$ and $\tau$ in Section \ref{sec:supplemental-gradient} of the supplement. 

The robust loss $\rho$  and its likelihood are
\begin{align}
\lossfun{x, \power, \sigma} &= {\abs{\power - 2} \over \power } \left( \left( {\left( \sfrac{x}{\sigma} \right)^2 \over \abs{\power - 2} } + 1 \right)^{\sfrac{\power}{2}} - 1 \right) \text{and} \\
\prob{\hat{y} \;|\; y, \power, \sigma} &= {\frac{1}{\sigma \partition{\power}}} \exp \left( -\lossfun{\hat{y} - y, \power, \sigma} \right),
\end{align}
with shape $\alpha \in [0, \infty)$, scale $\sigma \in (0, \infty)$, and normalization function $\partition{\alpha}$. 
This likelihood generalizes the normal, Cauchy, and Student's t distributions.

\section{Related Work}
\label{sec:related}

Likelihood optimization follows from maximum likelihood estimation \citep{hastie2009elements,bishop2006pattern}, yet is uncommon in practice for fitting deep regressors and classifiers for discriminative tasks.
However \cite{kendall2017uncertainties,kendall2018multi,barron2019loss,saxena2019data} optimize likelihood parameters to their advantage yet differ in their tasks, likelihoods, and parameterizations. 
In this work we aim to systematically experiment, clarify usage, and encourage their wider adoption.

Early work on regressing means and variances \citep{nix1994estimating} had the key insight that optimizing the full likelihood can fit these parameters and adapt the loss. 
Some recent works use likelihoods for loss adaptation, and interpret their parameters as the uncertainty \citep{kendall2017uncertainties,kendall2018multi}, robustness \citep{kendall2017uncertainties,barron2019loss,saxena2019data}, and curricula \citep{saxena2019data} of losses.
\cite{mackay1992bayesian} uses Bayesian evidence to select hyper-parameters and losses based on proper likelihood normalization.
\cite{barron2019loss} define a generalized robust regression loss, $\rho$, to jointly optimize the type and degree of robustness with global, data-independent, parameters.
\cite{kendall2017uncertainties} predict variances for regression and classification to handle data-dependent uncertainty.
\cite{kendall2018multi} balance multi-task loss weights by optimizing variances for regression and temperatures for classification.
These global parameters depend on the task but not the data, and are interpreted as inherent task uncertainty.
\cite{saxena2019data} define a differentiable curriculum for classification by assigning each training point its own temperature.
These data parameters depend on the index of the data but not its value.
We compare these different likelihood parametizations across tasks and distributions.

In the calibration literature, \cite{guo2017calibration} have found that deep networks are often miscalibrated, but they can be re-calibrated by cross-validating the temperature of the softmax. In this work we explore several generalizations of this concept. Alternatively, Platt scaling \citep{platt1999probabilistic} fits a sigmoid regressor to model predictions to calibrate probabilities. \cite{kuleshov2018accurate} re-calibrate regressors by fitting an Isotonic regressor to the empirical cumulative distribution function.

\section{Likelihood Parameter Types}
\label{sec:design}

We explore the space of likelihood parameter representations for model optimization and inference. Though we note that some losses, like adversarial losses, are difficult to represent as likelihoods, many different losses in the community have a natural probabilistic interpretation. Often, these probabilistic interpretations can be parametrized in a variety of ways. We explore two key axes of generality when building these loss functions: conditioning and dimensionality.

\begin{figure}
\centering
\begin{minipage}{.48\textwidth}
  \centering
  \includegraphics[width=\linewidth]{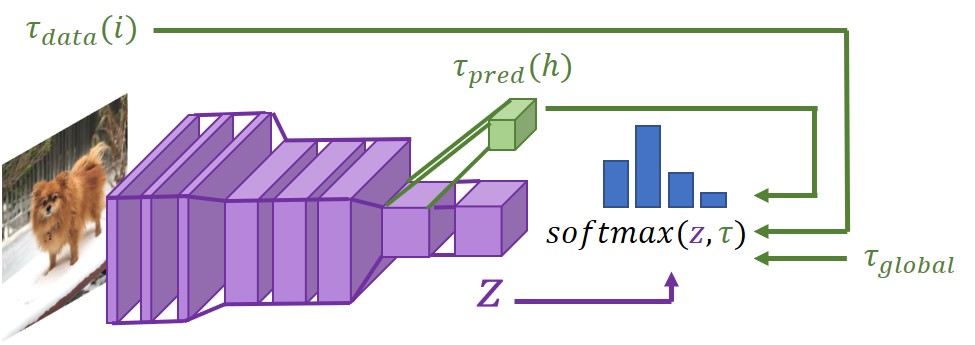}
  \captionof{figure}{An image classifier with three different types of temperature conditioning: global, model, and data. }
  \label{fig:conditioning}
\end{minipage}%
\hspace{.1in}
\begin{minipage}{.48\textwidth}
  \centering
  \includegraphics[width=\linewidth]{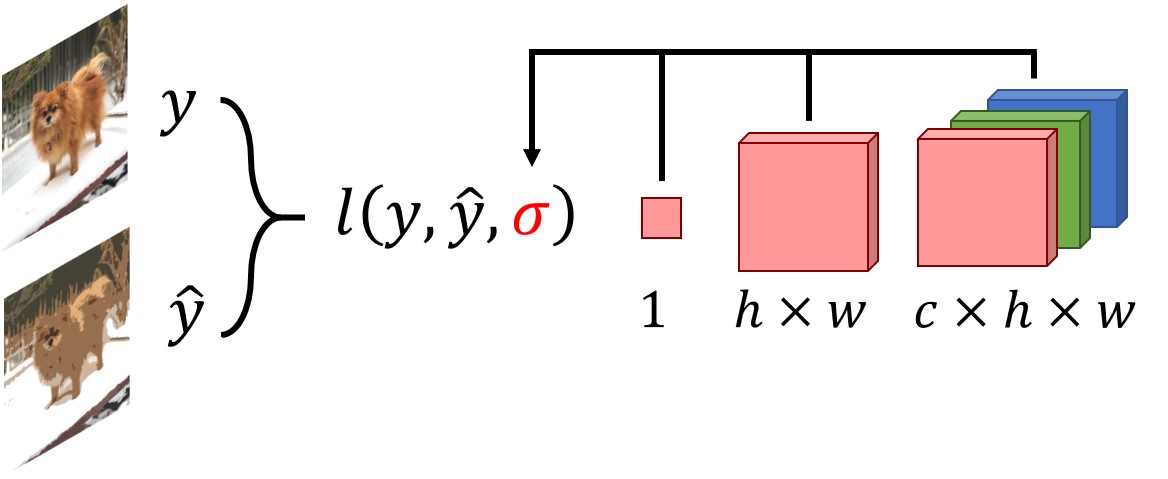}
  \captionof{figure}{An image loss function with three different likelihood parameter dimensionalities}
  \label{fig:dimensionality}
\end{minipage}
\end{figure}

\begin{figure}[t]
    \centering
    \includegraphics[width=\textwidth]{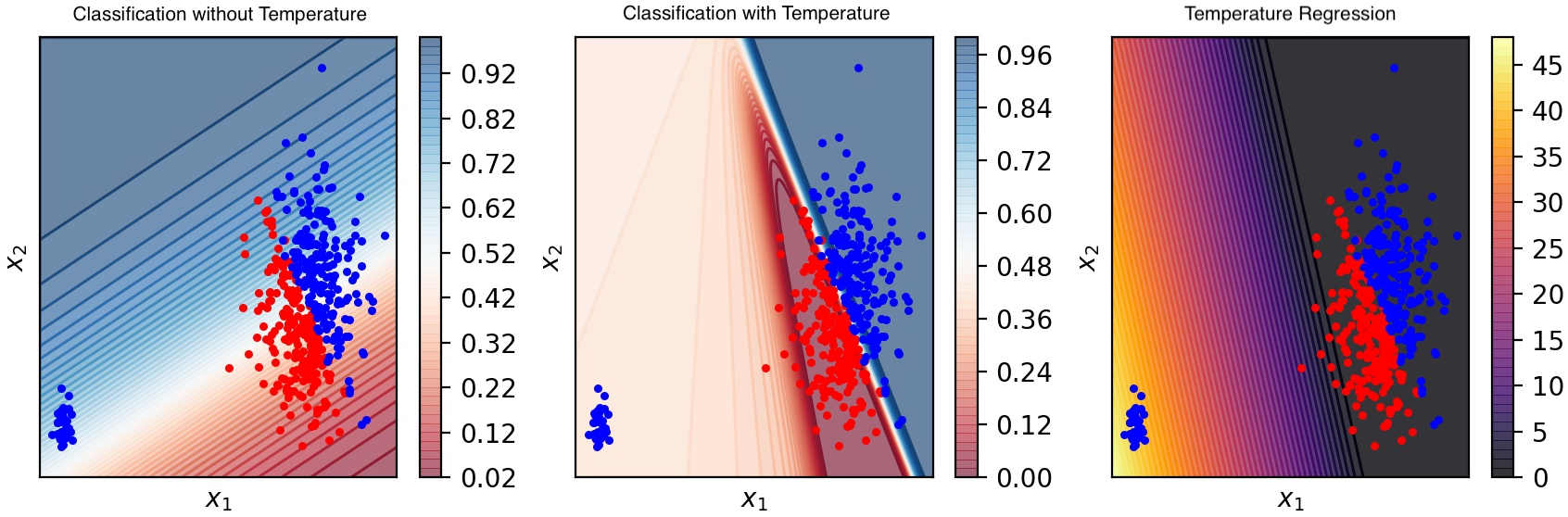}
    \vskip -0.1in
    \captionof{figure}{A synthetic logistic regression experiment. Regressing softmax temperature reduces the influence of outliers (blue, bottom-left), by locally raising temperature. The jointly optimized model achieves a more accurate classification.}
    \label{fig:robust_classification}
\end{figure}

\minisection{Conditioning}
We represent the likelihood parameters by three functional classes: global, data, and predicted.
\emph{Global} parameters, $\phi = c$, are independent of the data and model 
and define the same likelihood distribution for all points. 
\emph{Data} parameters, $\phi_i$, are conditioned on the index, $i$, of the data, $x_i$, but not its value.
Every training point is assigned an independent likelihood parameter, $\phi_i$ that define different likelihoods for each training point.
\emph{Predicted} parameters, $\phi(x) = g_\eta(x)$, are determined by a model, $g$, with parameters $\eta$ (not to be confused with the task model parameters $\theta$).
Global and predicted parameters can be used during training and testing, but data parameters are only assigned to each training point and are undefined for testing. We show a simple example of predicted temperature in Figure \ref{fig:robust_classification}, and an illustration of the parameter types in Figure \ref{fig:conditioning}.

We note that for certain global parameters like a learned Normal scale, changing the scale does not affect the optima, but does change the probabilistic interpretation. This invariance has led many authors to drop the scale from their formulations. However, when models can predict these scale parameters they can naturally remain calibrated in the presence of heteroskedasticity and outliers. Additionally we note that for the shape parameter of the robust likelihood, $\rho$, changing global parameters does affect model fitting. Previous works have adapted a global softmax temperature for model distillation \citep{hinton2015distilling}, and recalibration \citep{guo2017calibration}.

\minisection{Dimensionality}
The dimensionality, $|\phi|$, of likelihood parameters can vary with the dimension of the task prediction, $\hat{y}$.
For example, image regressors can use a single likelihood parameter for each image $|\phi| = 1$, RGB image channel $|\phi| = C$, or even every pixel $|\phi| = W \times H \times C$ as in Figure \ref{fig:dimensionality}. These choices correspond to different likelihood distribution classes. Dimensionality and Conditioning of likelihood parameters can interact. For example,  data parameters with $|\phi| = W \times H \times C$ would result in $N\times W \times H \times C$ additional parameters, where $N$ is the size of the dataset. This can complicate implementations and slow down optimization due to disk I/O when their size exceeds memory.
Table \ref{tab:constant-data-predicted} in the appendix contrasts the computational requirements of different likelihood parameter types. 

\section{Applications}
\label{sec:applications}

\subsection{Robust Modeling}
\label{sec:robust-modelling}

Data in the wild is noisy, and machine learning methods should be robust to noise, heteroskedasticity,  and corruption.
The standard mean squared error (MSE) loss is highly susceptible to outliers due to its fixed variance \citep{huber2004robust}.
Likelihood parameters naturally transform standard methods such as regressors, classifiers, and manifold learners into robust variants without expensive outer-loop of model fitting such as RANSAC \citep{fischler1981random} and Theil-Sen \citep{theil1992rank}. 
Figure \ref{fig:robust_classification} demonstrates this effect with a simple classification dataset, and we point readers to Figures \ref{fig:robust-models} and \ref{fig:robust_regression} of the Supplement for similar examples for regression and manifold learning.
By mitigating the harmful effect of outliers, regressing variance can often improve generalization ability in terms of MSE, accuracy, and NLL on an unseen test set. Furthermore, by allowing models to adjust their certainty in a principled manner, this also reduces the miscalibration (CAL) of regressors. Table \ref{table:robust-reg} shows this effect for deep models on the datasets used in \citep{kuleshov2018accurate}. 

Often the assumption of Gaussianity is too restrictive for regression models and one must consider more robust approaches. This has led many to investigate broader classes of likelihoods such as Generalized Linear Models (GLMs) \citep{nelder1972generalized} or the more recent general robust loss, $\rho$, of \cite{barron2019loss}.
To systematically explore how likelihood parameter dimension and conditioning affect model robustness and quality, we reproduce \cite{barron2019loss}'s variational auto-encoding \citep{kingma2014adam} (VAE) experiments on faces from the CelebA dataset \citep{liu2015faceattributes}. We compare global, predicted, and data parameters \citep{saxena2019data} as well as two natural choices of parameter dimensionality: a single set of parameters for the whole image, and a set of parameters for each pixel and channel. We note that predicted parameters achieve the best performance while maintaining speed and a small memory footprint. We visualize these parameters in Section \ref{sec:param-viz} of the Appendix, and show that they demarcate challenging areas of target images.
More details on experimental conditions, datasets, and models are provided in Sections \ref{sec:datasets} and \ref{sec:models} in the appendix. 

\begin{table}
    \centering
        \caption{MSE, Time, and Memory increase (compared to standard normal likelihood) for reconstruction by variational auto-encoders with different parameterizations of the robust loss, $\rho$. Predicted likelihood parameters yield more accurate reconstruction models.}
    \label{tab:vae}
    \begin{tabular}{ccccc}
        Param.  & Dim  & MSE & Time & Mem \\ \hline
        Global  & $1\!{\times}\!1\!{\times}\!1$   & 225.8 & \textbf{1.04$\times$} & \textbf{$<\!1$KB} \\
        Data      & ...                     & 244.2 & 2.70$\times$ & $0.6$GB\\ 
        Pred. & ...                     & 228.5 & \textbf{1.04$\times$} & $<\!1$MB\\
        Global  &  $H\!{\times}\!W\!{\times}\!C$  & 231.1 & 1.08$\times$ & $<\!1$MB\\ 
        Data      & ...                     & 252.6  & 9.42$\times$ & $4.4$GB \\
        Pred. & ...                     & \textbf{222.3} & 1.08$\times$ & $<\!1$MB \\
    \end{tabular}
\end{table}

\setlength{\tabcolsep}{3pt}
\begin{table}[t]
\caption{Effect of predicted likelihood parameters on Calibration (CAL) \citep{kuleshov2018accurate}, MSE, and NLL evaluated on test data for deep regressors. Linear results are in Table \ref{table:robust-reg-linear} of the supplement.}
\label{table:robust-reg}

\vskip 0.15in
\begin{center}
\begin{small}
\begin{sc}
\begin{tabular}{c|cc|cc|cc}
        &  \multicolumn{2}{c|}{CAL}               & \multicolumn{2}{c|}{MSE}            & \multicolumn{2}{c}{NLL}     \\
 Dataset    & Base           & Temp           & Base  & Temp           & Base             & Temp            \\ \hline
 crime      & \textbf{0.018} & 0.146          & 0.028 & \textbf{0.088} & \textbf{220.1} & 478.4         \\
 kin. & 0.122          & \textbf{0.001} & 0.064 & \textbf{0.006} & 0.061            & \textbf{-0.471} \\
 bank       & 0.417          & \textbf{0.001} & 0.127 & \textbf{0.008} & 1.361            & \textbf{-1.464} \\
 wine       & 0.023          & \textbf{0.003} & 0.032 & \textbf{0.011} & 1.614            & \textbf{0.302}  \\
 mpg        & 0.208          & \textbf{0.006} & 0.083 & \textbf{0.020} & \textbf{0.307}   & 5.050           \\
 cpu        & 0.554          & \textbf{0.021} & 0.150 & \textbf{0.022} & \textbf{-0.102}  & 12.218          \\
 soil       & 0.602          & \textbf{0.307} & 0.160 & \textbf{0.100} & -0.131           & \textbf{-4.078} \\
 fried      & 0.472          & \textbf{0.000} & 0.129 & \textbf{0.002} & 0.301            & \textbf{-1.039} \\ 
\end{tabular}
\end{sc}
\end{small}
\end{center}
\end{table}

\subsection{Outlier Detection}

\begin{figure}[t]
\centering
\includegraphics[width=\textwidth]{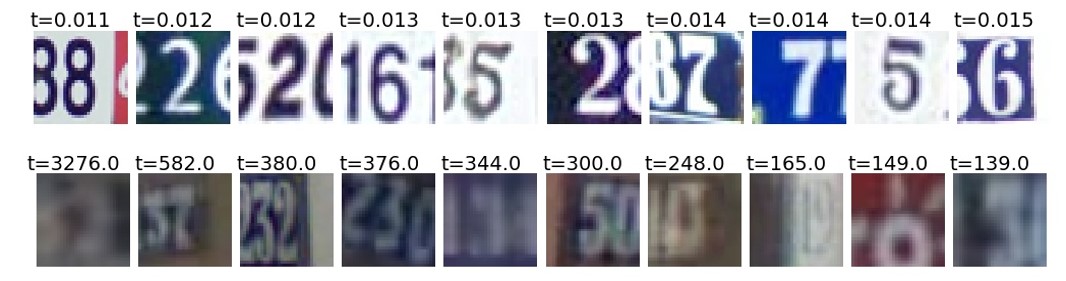}
\caption{
The data with the lowest (top) and highest (bottom) predicted temperatures in the SVHN dataset.
High temperature entries are blurry, cropped poorly, and generally difficult to classify.
}
\label{fig:bad_data}
\end{figure}

Likelihood parameter prediction gives models a direct channel to express their ``uncertainty'' for each data-point with respect to the task. This allows models to naturally down-weight and clean outliers from the dataset which can improve model robustness. Consequently, one can harness this effect to create outlier detectors from \textit{any} underlying model architecture by using learned scales or temperatures as an outlier score function. Furthermore, predicted likelihood parameters allow these methods to detect outliers in unseen data. In Figure \ref{fig:bad_data} we show how auditing temperature or noise parameters can help practitioners spot erroneous labels and poor quality examples. In particular, the temperature of an image classifier correlates strongly with blurry, dark, and difficult examples on the Street View House Number (SVHN) dataset. 

\begin{figure}[t]
    \centering
    \begin{minipage}[t]{0.49\textwidth}
        \vspace{0pt}
        \captionof{table}{Median outlier detection performance of several methods across 22 benchmark datasets from ODDS. }
        \label{tab:outlier-det}
        \begin{center}
        \vspace{-.1in}
        \begin{tabular}{cc}
        Method       & Median AUC \\ \hline
        LOF          & .669      \\
        FB           & .702      \\
        ABOD         & .727      \\
        AE           & .737      \\
        VAE          & .792      \\
        COPOD        & .799      \\
        PCA          & .808      \\
        OCSVM        & .814      \\
        MCD          & .820      \\
        KNN          & .822      \\
        HBOS         & .822      \\
        IF           & .823      \\
        CBLOF        & .836      \\
        AE+S (Ours)  & .846      \\
        \textbf{PCA+S (Ours)} & \textbf{.868}     
        \end{tabular}
        \end{center}
    \end{minipage}
    \hfill
    \begin{minipage}[t]{0.49\textwidth}
    \centering
    \vspace{0pt}

    \includegraphics[width=\linewidth]{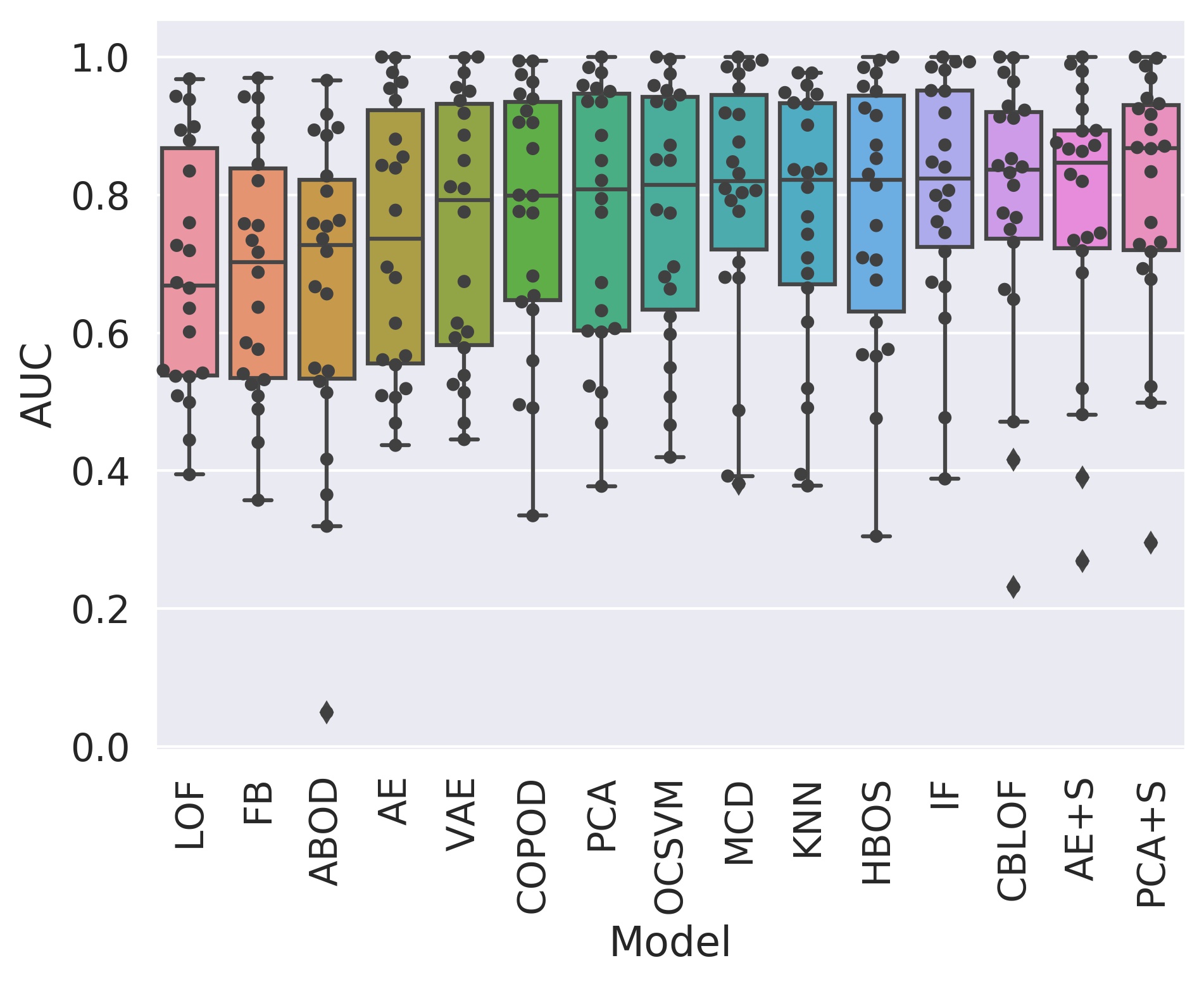}
    \caption{Distribution of Outlier Detection AUC across the ODDS Benchmark. Our approaches, PCA+S and AE+S, are competitive with other Outlier Detection systems.}
    \label{fig:my_label}
    \end{minipage}
\end{figure}

One can also leverage auto-encoders \citep{kramer1991nonlinear} or self-supervision \citep{zhang2016colorful, tian2019contrastive} to yield label-free measures of uncertainty. We use this approach to create a simple outlier detection algorithms by considering deep (AE+S) and linear (PCA+S) auto-encoders with a data-conditioned scale parameters as outlier scores. In Table \ref{tab:outlier-det} we quantitatively demonstrate the quality of these simple likelihood parameter approaches across 22 datasets from the Outlier Detection Datasets (ODDS), a standard outlier detection benchmark \citep{rayana2016odds}. The ODDS benchmark supplies ground truth outlier labels for each dataset, which allows one to treat outlier detection as an unsupervised classification problem. We compare against a variety of established outlier detection approaches including: One-Class SVMs (OCSVM) \citep{scholkopf2000support}, Local Outlier Fraction (LOF) \citep{breunig2000lof}, Angle Based Outlier Detection (ABOD) \citep{kriegel2008angle}, Feature Bagging (FB) \citep{lazarevic2005feature}, Auto Encoder Distance (AE) \citep{aggarwal2015outlier}, K-Nearest Neighbors (KNN) \citep{ramaswamy2000efficient,angiulli2002fast}, Copula Based Outlier Detection (COPOD) \citep{li2020copod}, Variational Auto Encoders (VAE) \citep{kingma2013auto},  Minimum Covariance Determinants with Mahlanohbis Distance (MCD) \citep{rousseeuw1999fast,hardin2004outlier}, Histogram-based Outlier Scores (HBOS) \citep{goldstein2012histogram}, Principal Component Analysis (PCA) \citep{shyu2003novel}, Isolation Forests (IF) \citep{liu2008isolation,liu2012isolation}, and the Clustering-Based Local Outlier Factor (CBLOF) \citep{he2003discovering}. For experimental details please see Sections \ref{sec:datasets} and \ref{sec:models} of the Appendix.

Our methods (PCA+S and AE+S) use a similar principle as isolation-based approaches that determine outliers based on how difficult they are to model. In existing approaches, outliers influence and skew the isolation model which causes the model to exhibit less confidence on the whole. This hurts a model's ability to distinguish between inliers and outliers. In contrast, our approach allows the underlying model to down-weight outliers. This yields a more consistent model with a clearer decision boundary between outliers and inliers as shown in Figure \ref{fig:robust_classification}. As a future direction of investigation we note that our approach is model-architecture agnostic, and can be combined with domain-specific architectures to create outlier detection methods tailored to images, text, and audio. 

\setlength{\tabcolsep}{2pt}

\subsection{Adaptive Regularization with Prior Parameters}

\begin{figure}[t]
    \centering
    \begin{minipage}[t]{0.32\textwidth}
    \includegraphics[width=\linewidth]{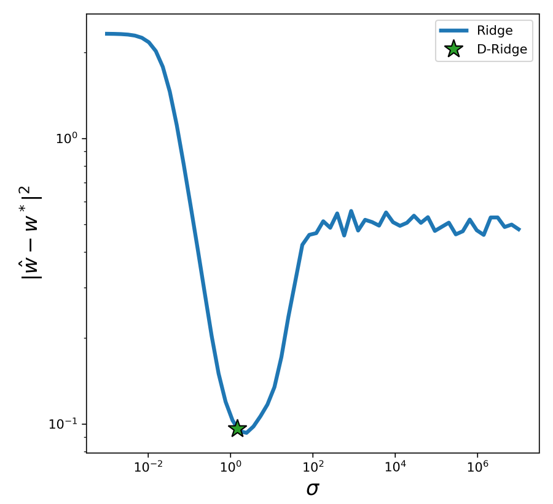}
    \end{minipage}
    \hfill
    \begin{minipage}[t]{0.32\textwidth}
    \includegraphics[width=.92\linewidth]{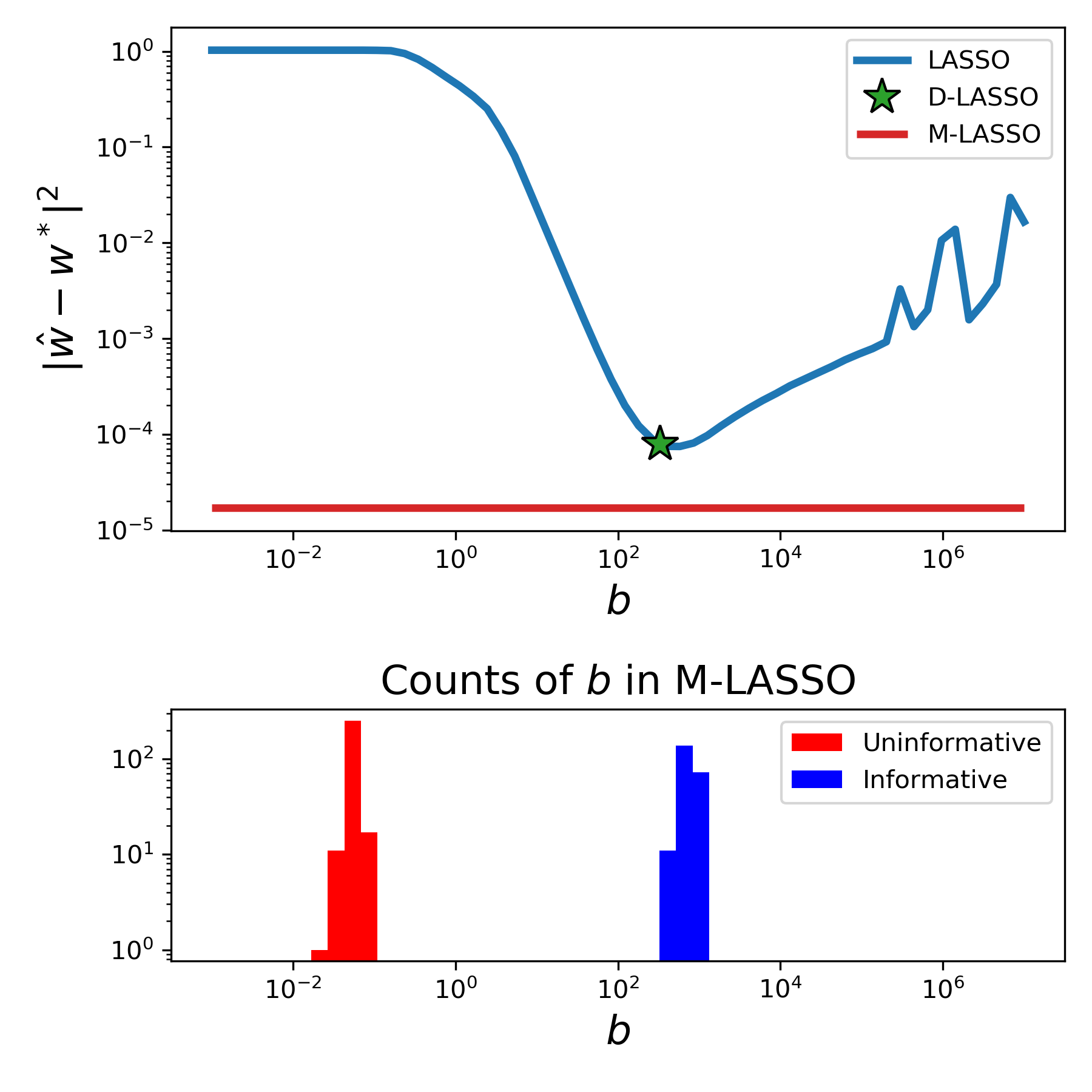}
    \end{minipage}
    \hfill
    \begin{minipage}[t]{0.32\textwidth}
    \includegraphics[width=.92\linewidth]{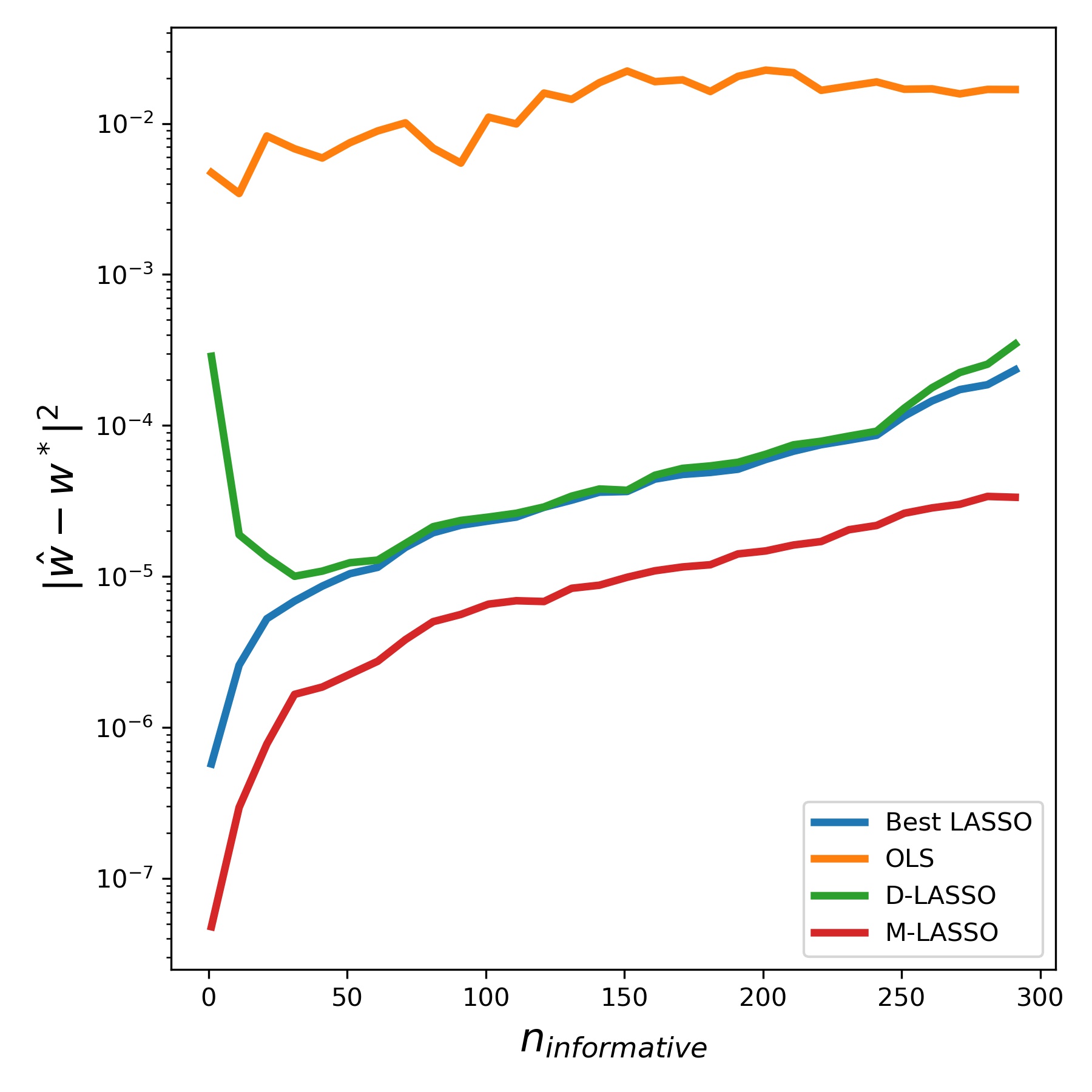}
    \end{minipage}
    \caption{Performance of $L2$ (left) and $L1$ (middle) regularized linear regression on a 500 dimensional synthetic dataset where the true parameters, $w^*$, are known. Dynamic Ridge (D-Ridge) and D-LASSO regression find the regularization strength that best estimates the true parameters. M-LASSO outperforms any single global regularization strength and does not shrink informative weights. (right) Performance of adaptive $L1$ regularization methods as a function of true model sparsity. In all cases, Multi-LASSO outperforms other methods by orders of magnitude.}
    \label{fig:adaptive_reg}
\end{figure}

In addition to optimizing the shape and scale of the likelihood distribution we can use the same approach to optimize a model's \textit{prior} distribution. More specifically, we propose adaptive regularizers for a model's parameters, $\theta$. This approach optimizes the distribution parameters of the prior, $\phi_{\text{prior}}$, to naturally tune the degree of regularization.
In particular, the Normal (Ridge, L2) and Laplace (LASSO, L1) priors, with scale parameters $\sigma$ and $b$, regularize model parameters for small magnitude and sparsity respectively \citep{hastie2009elements}.
The degree of regularization, $\lambda \in [0, \infty)$, is conventionally a hyperparameter of the regularized loss function:
\begin{equation}
\min_\theta \sum_i^N (\hat{y_i} := f_\theta(x_i) - y_i)^2 + \lambda \sum_j^P |\theta_j|.
\end{equation}
We note that we cannot choose $\lambda$ by direct minimization because it admits a trivial minimum at $\lambda = 0$.
In the linear case, one can select this weight efficiently using Least Angle Regression \citep{efron2004least}.
However, in general $\lambda$ is usually learned through expensive cross validation methods. Instead, we retain the prior with its scale parameter, and jointly optimize over the full likelihood:

\begin{equation}
\min_{\theta,\sigma,b} \sum_i^N 
\left(\frac{1}{2\sigma^2}(\hat{y}_i - y_i)^2 + \log\sigma\right) + \sum_j^P \left(\frac{|\theta_j|}{b} + \log b\right)
\end{equation}

This approach, the Dynamic Lasso (D-LASSO), admits no trivial solution for the prior parameter $b$, and must balance the effective regularization strength, $\frac{1}{b}$, with the normalization factor, $\log b$.
D-LASSO selects the degree of regularization by gradient descent, rather than expensive black-box search.
In Figure \ref{fig:adaptive_reg} (left) and (middle) we show that this approach, and its Ridge equivalent, yield ideal settings of the regularization strength. 
Figure \ref{fig:adaptive_reg} (right) shows D-LASSO converges to the best LASSO regularization strength for a variety of true-model sparsities. 
As a further extension, we replace the global $\sigma$ or $b$ with a $\sigma_j$ or $b_j$ for each model parameter, $\theta_j$, to locally adapt regularization to each model weight (\textbf{M}ulti-Lasso). This consistently outperforms any global setting of the regularization strength and shields important weights from undue shrinkage \ref{fig:adaptive_reg} (middle).

\subsection{Re-calibration}
 
The work of \citep{guo2017calibration} shows that modern networks are accurate, yet systematically overconfident, a phenomenon called mis-calibration.
We investigate the role of optimizing likelihood parameters to re-calibrate models. More specifically, we can fit likelihood parameter regressors on a validation set to modify an existing model's confidence to better align with the validation set. This approach is a generalization of \cite{guo2017calibration}'s Temperature Scaling method, which we refer to as Global Scaling (GS) for notational consistency. Global Scaling re-calibrates classifiers with a learned global parameter, $\tau$ in the loss function: $\sigma(\vec{x},\tau)$.

Fitting model-conditioned likelihood parameters to to a validation set defines a broad class of re-calibration strategies. From these we introduce three new re-calibration methods. Linear Scaling (LS) learns a linear mapping, $l$, to transform logits to a softmax temperature: $\sigma(\vec{x},l(\vec{x}))$. Linear Feature Scaling (LFS) learns a linear mapping, $l$, to transform the \textit{features} prior to the logits, $\vec{f}$, to a softmax temperature: $\sigma(\vec{x},l(\vec{f}))$. Finally, we introduce Deep Scaling (DS) for regressors which learns a nonlinear network, $N$, to transform features, $\vec{f}$, into a temperature: $\sigma(\vec{x},N(\vec{f}))$. 

In Table \ref{tab:class-calibration} we compare our recalibration approaches to the previous state of the art: Global Scaling. We note that \citep{guo2017calibration} have already shown that Global Scaling outperform Bayesian Binning into Quantiles \citep{naeini2015obtaining}, Histogram binning \citep{zadrozny2001obtaining}, and Isotonic Regression. We recalibrate both ResNet50 \citep{he2016deep} and DenseNet121 \citep{huang2017densely} on a variety of vision datasets. We measure classifier miscalibration using the Expected Calibration Error (ECE) \citep{guo2017calibration} to align with prior art. We additionally evaluate Isotonic recalibration, Platt Scaling \citep{platt1999probabilistic}, and Vector Scaling (VS) \citep{guo2017calibration}, which learns a vector, $\vec{v}$, to re-weight logits: $\sigma(\vec{v} \vec{x}, 1)$. LS and LFS tend to outperform other approaches like GS and VS, which demonstrates that richer likelihood parametrizations can improve calibration akin to how richer models can improve prediction.

For recalibrating regressors, we compare against the previous state of the art, \cite{kuleshov2018accurate}, who use an Isotonic regressor to correct a regressors' confidence. We use the same datasets, and regressor calibration metric (CAL) as \cite{kuleshov2018accurate}.
Table \ref{tab:reg-calibration} shows that our approaches can outperform this baseline as well as the regression equivalent of Global Scaling. 

\setlength{\tabcolsep}{4pt}
\begin{table}[t]
\caption{Comparison of calibration methods by ECE for ResNet-50 (RN50) and DenseNet-121 (DN121) architectures on test data.
Our predicted likelihood parameter methods: Linear Scaling (LS) and Linear Feature Scaling (LFS) outperform other approaches.
In all cases our methods reduce miscalibration with comparable computation time as GS.}
\label{tab:class-calibration}
\begin{center}
\begin{tabular}{ccccccccc}
Model    & Dataset  & Uncalibrated  & Platt & Isotonic & GS    & VS    & LS    & LFS   \\ \hline
RN50   & CIFAR-10  & .250 & .034 &.053 & .046 & .037 & \textbf{.018} & \textbf{.018} \\
RN50   & CIFAR-100 & .642 & .061 & .072 & .035 & .044 & \textbf{.030} & .173 \\
RN50   & SVHN     & .072 & .053 &.010 & .029 & .022 & \textbf{.009} & \textbf{.009} \\
RN50   & ImageNet & .430 & .018 & .070 & .019 & .023 & .026 & \textbf{.015} \\
DN121 & CIFAR-10  & .253 & .048 & .042 & .039 & .034 & \textbf{.028} & \textbf{.028} \\
DN121 & CIFAR-100 & .537 &.049 & .067 & .024 & .024 & \textbf{.014} & .031 \\
DN121 & SVHN     & .079 & .018 & \textbf{.010} & .022 & .017 & .011 & \textbf{.010} \\
DN121 & ImageNet & .229 & .028 & .095 & .021 & \textbf{.019} & .043 & \textbf{.019} \\ 
\end{tabular}
\end{center}
\end{table}

\setlength{\tabcolsep}{2pt}
\begin{table}[h]
\caption{Comparison of regression calibration methods as evaluated by their calibration error as defined in \citep{kuleshov2018accurate}. Predicted likelihood parameters often outperform other methods.
}
\label{tab:reg-calibration}
\begin{center}
\begin{tabular}{cccccc}

 Dataset    & Uncalibrated    & Isotonic        & GS              & LS              & DS              \\ \hline
crime      & 0.3624          & 0.3499          & 0.0693          & \textbf{0.0125} & 0.0310          \\
 kinematics & 0.0164          & 0.0103          & 0.0022          & \textbf{0.0021} & 0.0032          \\
 bank       & 0.0122          & 0.0056          & 0.0027          & 0.0024          & \textbf{0.0020} \\
 wine       & 0.0091          & 0.0108          & 0.0152          & 0.0131          & \textbf{0.0064} \\
 mpg        & 0.2153          & 0.2200          & 0.1964          & 0.1483          & \textbf{0.0233} \\
 cpu        & 0.0862          & \textbf{0.0340} & 0.3018          & 0.2078          & 0.1740          \\
 soil       & 0.3083          & \textbf{0.3000} & 0.3130          & 0.3175          & 0.3137          \\
 fried      & 0.0006          & \textbf{0.0002} & \textbf{0.0002} & \textbf{0.0002} & \textbf{0.0002} \\
\end{tabular}
\end{center}
\end{table}


\section{Conclusion}

Optimizing the full likelihood can improve model quality by adapting losses and regularizers.
Full likelihoods are agnostic to the architecture, optimizer, and task, which makes them simple substitutes for standard losses. 
Global, data, and predicted likelihood parameters offer different degrees of expressivity and efficiency.
In particular, predicted parameters adapt the likelihood to each data point during training and testing without significant time and space overhead.
By including these parameters in a loss function one can improve a model's robustness and generalization ability and create new classes of outlier detectors and recalibrators that outperform baselines.
More generally, we hope this work encourages joint optimization of model and likelihood parameters, and argue it is likely that your loss should be a likelihood.


\bibliography{iclr2021_conference}
\bibliographystyle{iclr2021_conference}

\appendix
\onecolumn
\section*{Appendix}
\section{Gradient Optimization of Variance and Temperature}
\label{sec:supplemental-gradient} 

For completeness, we state the derivative of the normal NLL with respect to the variance \eqref{eq:grad-sigma} and the derivative of the softmax NLL with respect to the temperature \eqref{eq:grad-tau}.
The normal NLL is well-known, and its gradient w.r.t. the variance $\sigma$ was first used to fit networks for heteroskedastic regression \citep{nix1994estimating} and mixture modeling \citep{bishop1994mixture}.
The softmax with temperature is less widely appreciated, and we are not aware of a reference for its gradient w.r.t. the temperature.
For this derivative, recall that the softmax is the gradient of $\log \sum \exp$, and see \eqref{eq:like-softmax}.

\begin{equation}
\label{eq:grad-sigma}
\frac{\partial \nll_{\normal}}{\partial \sigma^2} = \frac{1}{2\sigma^2} \Big(-1 + \frac{1}{\sigma^2} (\hat{y} - y)^{2} \Big).
\end{equation}

\begin{equation}
\label{eq:grad-tau}
\frac{\partial \nll_{\softmax}}{\partial \tau} = -z_y + \sum_c z_c \frac{e^{z_c\tau}}{\sum_{c'}e^{z_{c'}\tau}}.
\end{equation}




\section{Understanding Where Uncertainty is Modelled}

\begin{figure}[h]
  \centering
    \caption{Area Under PR Curve of Softmax Temperature as a data corruption classifier on CIFAR10 with synthetic corruptions. Data temperatures (DT) train slower than model temperatures (MT), hence some  uncertainty is modelled by $f_{theta}$.}
    \includegraphics[width=.45\textwidth]{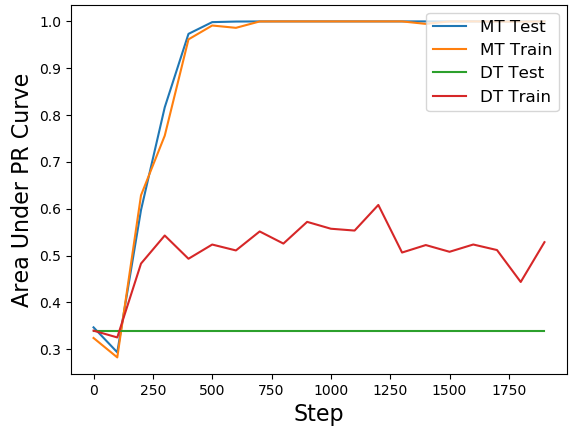}
    \label{fig:outlier_aupr}
\end{figure}

For classifiers with a parametrized temperature, models have the ``choice'' to store uncertainty information in either the model $f_{\theta}$ that can reduce the size of the logit vector, $z$, or the likelihood parameters $\phi$ which can scale the temperature parameter, $\tau$. Note that for regressors, this information can only be stored in $\phi$. The fact that uncertainty information is split between the model and likelihood can sometimes make it difficult to interpret temperature as the sole detector of outliers. In particular, if the uncertainty parameters, $\phi$, train slower than the model parameters, the network might find it advantageous to move critical uncertainty information to the model. This effect is illustrated by Figure \ref{fig:outlier_aupr}, which shows that with data parameters on a large dataset, the uncertainty required to detect outliers moves into the model.

\newpage
\section{Visualizing Robustness}

\begin{figure}[h]
    \centering
    \begin{minipage}[t]{0.32\textwidth}
    \includegraphics[width=\linewidth]{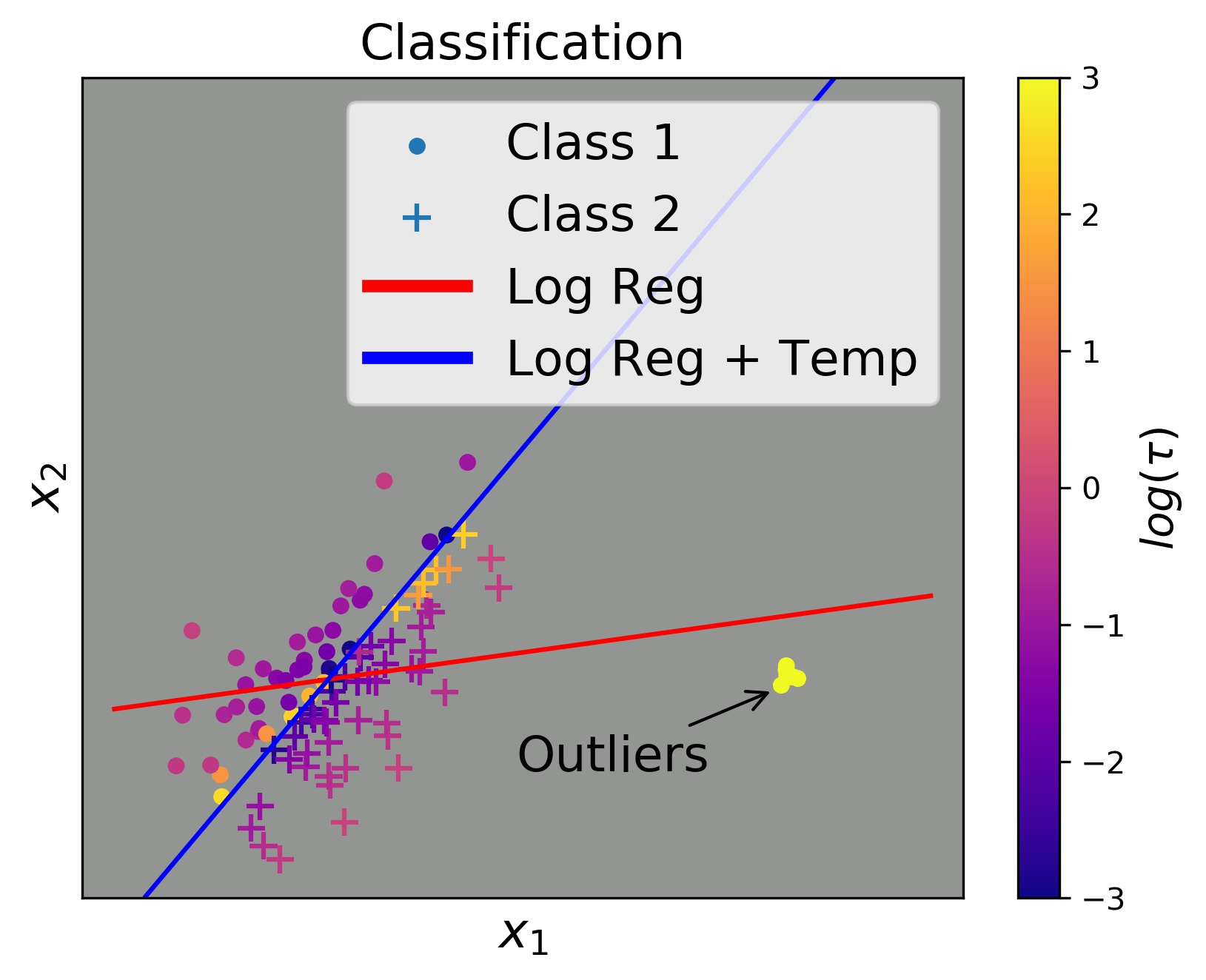}
    \end{minipage}
    \hfill
    \begin{minipage}[t]{0.32\textwidth}
    \includegraphics[width=\linewidth]{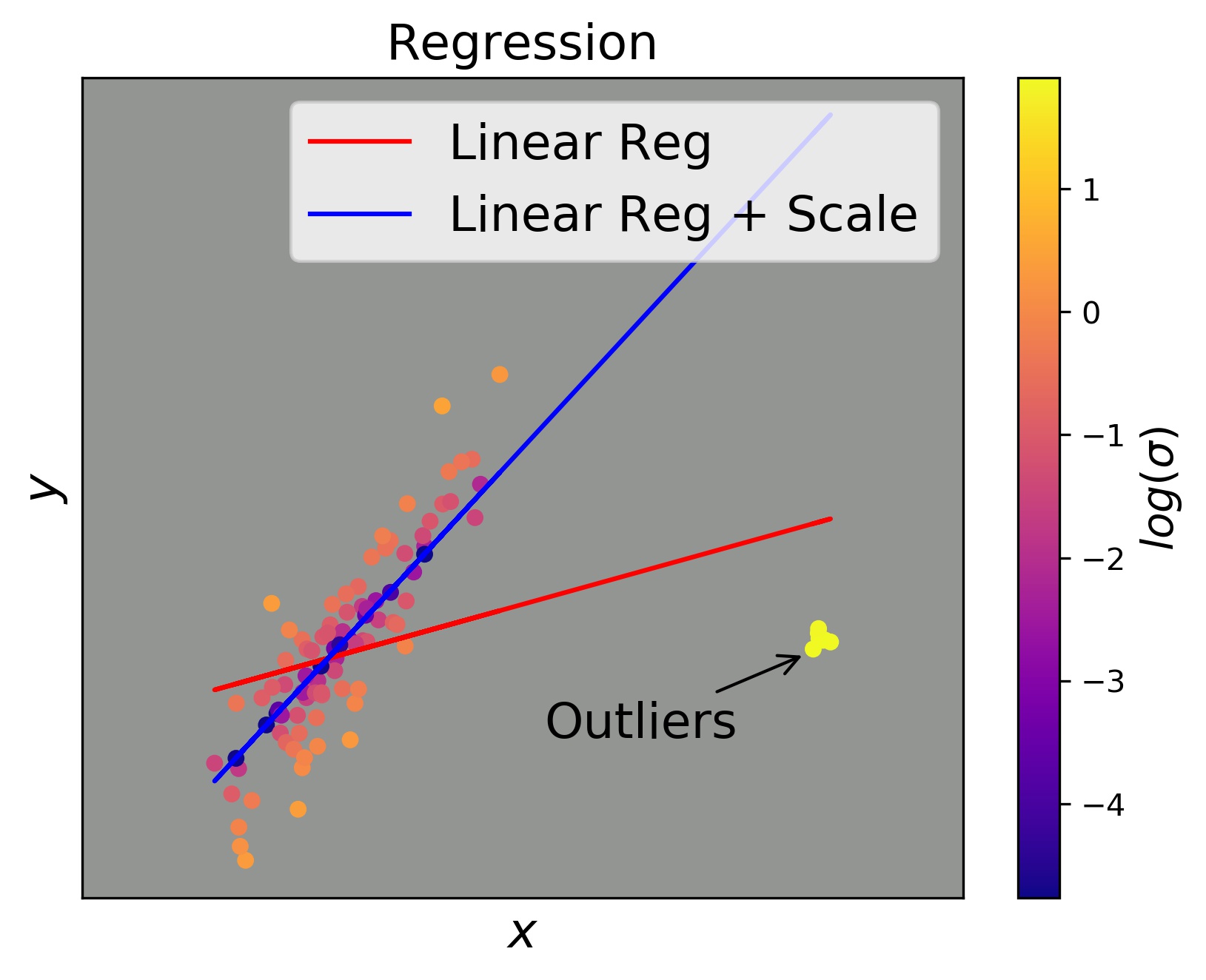}
    \end{minipage}
    \hfill
    \begin{minipage}[t]{0.32\textwidth}
    \includegraphics[width=\linewidth]{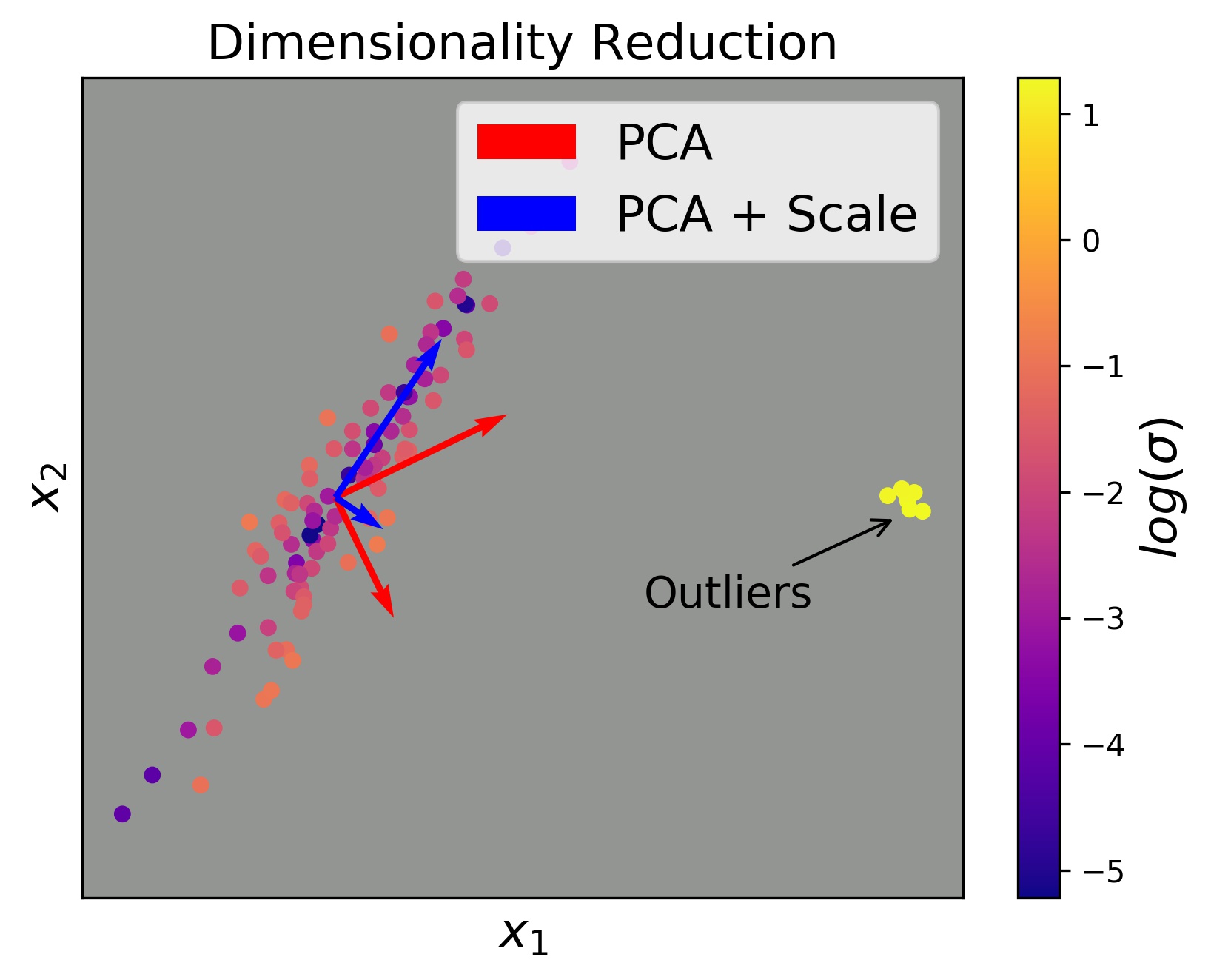}
    \end{minipage}
    \caption{The effect of likelihood parameters across three domains: Classification, Regression, and Dimensionality Reduction. In each, adding likelihood parameters yield robust variants of the original approach independent of the model architecture or task.}
    \label{fig:robust-models}
\end{figure}

Likelihood parameters can apply in a variety of different modelling domains such as regression, classification, and dimensionality reduction. Figure \ref{fig:robust-models} shows the beneficial effects of these parameters across these domains. 

\newpage
\section{Robust Tabular Regression}

We note that the same approaches of \ref{sec:robust-modelling} also apply to linear and deep regressors. Tables \ref{table:robust-reg-linear} and \ref{table:robust-reg} show these findings respectively.

\setlength{\tabcolsep}{3pt}
\begin{table}[h]
\caption{Effect of model based likeihood parameters on  Calibration (CAL) \citep{kuleshov2018accurate}, MSE and NLL evaluated on unseen data for in linear regressors.}
\label{table:robust-reg-linear}
\begin{center}
\begin{small}
\begin{sc}
\begin{tabular}{c|cc|cc|cc}
            &  \multicolumn{2}{c|}{CAL}               & \multicolumn{2}{c|}{MSE}            & \multicolumn{2}{c}{NLL}     \\
 Dataset    & Base           & Temp           & Base  & Temp           & Base             & Temp            \\ \hline

 crime      & 0.097          & \textbf{0.007} & 0.057 & \textbf{0.020} & \textbf{0.528}   & 17.497          \\
 kin. & 0.022          & \textbf{0.007} & 0.033 & \textbf{0.016} & 4.749            & \textbf{0.227}  \\
 bank       & 0.273          & \textbf{0.002} & 0.095 & \textbf{0.009} & 0.225            & \textbf{-1.045} \\
 wine       & 0.013          & \textbf{0.002} & 0.031 & \textbf{0.010} & 7.707            & \textbf{0.332}  \\
 mpg        & 0.093          & \textbf{0.016} & 0.057 & \textbf{0.030} & 0.217            & \textbf{-0.183} \\
 cpu        & 0.359          & \textbf{0.129} & 0.111 & \textbf{0.044} & 0.390            & \textbf{-2.812} \\
 soil       & 0.122          & \textbf{0.023} & 0.064 & \textbf{0.026} & 1.309            & \textbf{-2.427} \\
 fried      & 0.077          & \textbf{0.001} & 0.051 & \textbf{0.006} & 2.624            & \textbf{-0.195} \\
\end{tabular}
\end{sc}
\end{small}
\end{center}
\end{table}

\begin{figure}[h]
    \centering
    \includegraphics[width=\linewidth]{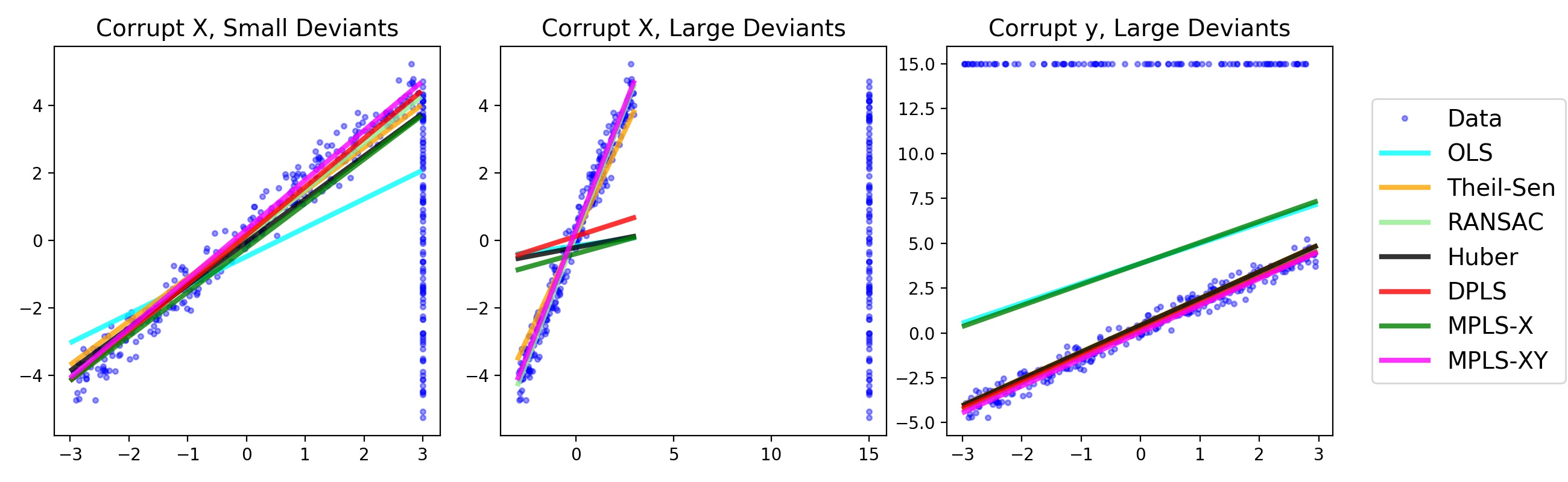}
    \caption{Comparison of robust regression methods. The choice of likelihood parameter conditioner has a significant effect on the inductive bias of outlier detection. Data Parameter Least Squares DPLS does not yield the appropriate inductive bias in this particular large X deviant dataset. The XY conditioned Model Parameters, MPLS-XY, induce a helpful bias that rivals the performance of RANSAC at a fraction of the fitting time}
    \label{fig:robust_regression}
\end{figure}

In addition to investigating how parametrized likelihoods affect deep models, we also perform the same comparisons and experiments for linear models. In this domain, we find that the same properties still hold, and because of the limited number of parameters, these models often benefit significantly more from likelihood parameters. In Figure \ref{fig:robust_regression} we show that parametrizing Gaussian scale leads to similar robustness properties as the Theil-Sen, RANSAC, and Huber methods at a fraction of the computational cost. Furthermore, we note that likihood parameters could also be applied to Huber regression to adjust its scale as is the case for Normal variance, $\sigma$. In Table \ref{table:robust-reg-linear} we show that adding a learned normal scale regressor to linear regression can improve calibration and generalization. 

\newpage
\section{Visualizing Predicted Shape and Scale for Auto-Encoding}
\label{sec:param-viz}

\begin{figure}[h]
\centering
\includegraphics[width=.8\linewidth]{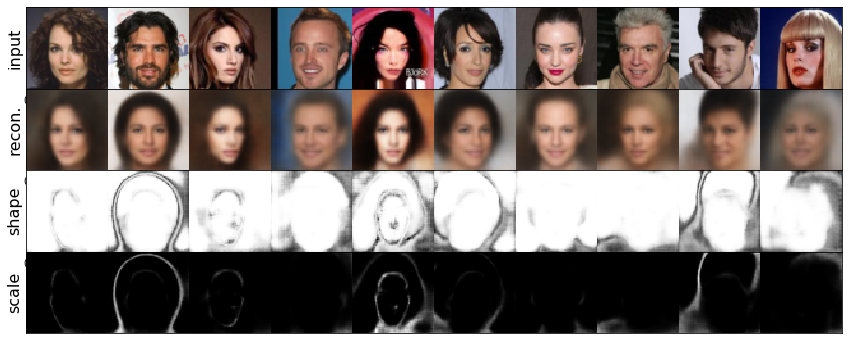}
\includegraphics[width=.8\linewidth]{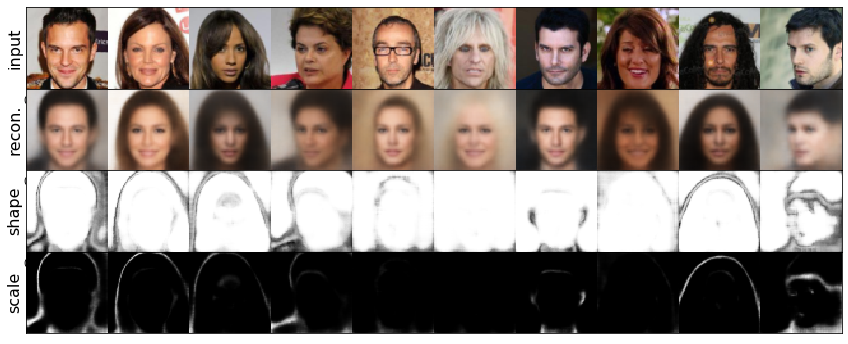}
\includegraphics[width=.8\linewidth]{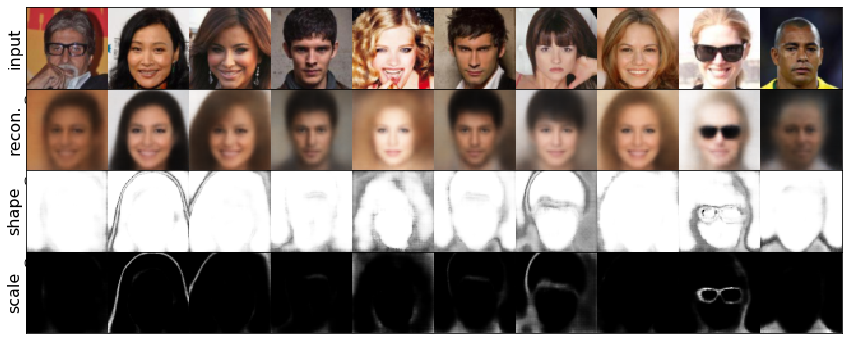}
\caption{
With predicted likelihood parameters, the shape and scale of the distribution (and hence loss) is adapted to each pixel by regression from the input.
The input, its reconstruction, and the predicted shape ($\alpha$) and scale ($\sigma$) are shown on CelebA validation images.
The likelihood parameters do indeed vary with the input images.
Across the whole validation set, the predicted shape is generally in the range $[1.8, 3.0]$ and the predicted scale is generally in the range $[0.20, 0.43]$.
}
\label{fig:vae-shape-scale}
\end{figure}

In Section \ref{sec:robust-modelling} we experiment with variational auto-encoding by the generalized robust loss $\rho$.
The likelihood corresponding to $\rho$ has parameters for shape ($\alpha$) and scale ($\sigma$).
When these parameters are predicted, by regressing them as part of the model, they can vary with the input to locally adapt the loss.
In Figure \ref{fig:vae-shape-scale} we visualize the regressed shape and scale of each pixel for images in the CelebA validation set.
Note that only predicted likelihood parameters can vary in this way, since data parameters are not defined during testing.

\section{Deep Regressors Miscalibrate}

The work of \cite{kuleshov2018accurate} establishes Isotonic regression as a natural baseline for regressors. 
In our experiments on classifiers in Table \ref{tab:class-calibration}, and on regressors in table \ref{tab:reg-calibration}, we found that additional likelihood modeling capacity for temperature and scale was beneficial for re-calibration across several datasets.
This demonstrates that the space of deep network calibration is rich, but with an inductive bias towards likelihood parametrization.
We also find that deep regressors suffer from the same over-confidence issue that plagues deep classifiers \citep{guo2017calibration}.
Figure \ref{fig:regressor_calibration_by_size} in the supplement shows this effect. More details on experimental conditions, datasets, and models are provided in sections \ref{sec:datasets} and \ref{sec:models} in the Supplement. 

We confirm an that deep classifiers miscalibrate as a function of layer size, which is analogous to the results for classifiers found by  \cite{guo2017calibration}. We plot regression calibration error as a function of network layer size for a simple single layer deep network on a synthetic regression dataset. As layer size grows the network has the capacity to over-fit the training data and hence under-estimate's its own errors. We note that this effect appears on the other datasets reported, and across other network architectures. We posit that overconfidence will occur with any likelihood distribution as this is a symptom of over-fitting.

\begin{figure}[h]
    \centering
    \includegraphics[width=.5\linewidth]{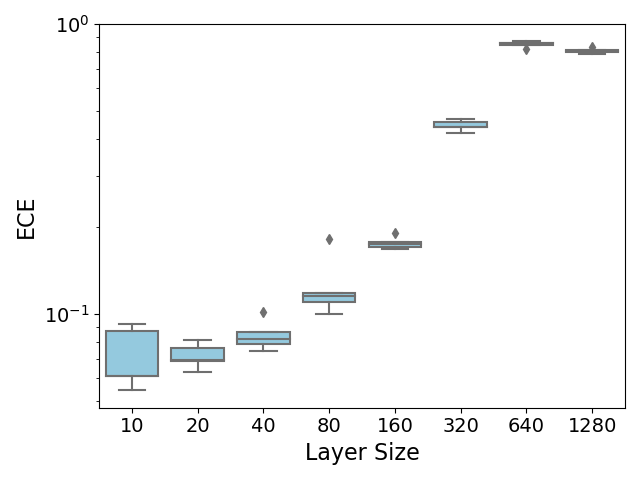}
    \caption{Regression Calibration Error as defined in \cite{kuleshov2018accurate} by network layer size. Experiments performed on a synthetic dataset with a 1 layer deep ReLU network. This shows regression overconfidence and miscalibration are a function of a network's ability to over-fit on data. A finding that agrees with \cite{guo2017calibration}.}
    \label{fig:regressor_calibration_by_size}
\end{figure}






\newpage
\section{Improving Optimization Stability}


When modelling Softmax temperatures and Gaussian scales with dedicated networks we encountered significant instability due to exploding gradients at low temperature scales and vanishing gradients at high temperature scales.
We discovered that it was desire-able to have a function mapping $\mathbb{R} \to \mathbb{R}^{>0}$ that was smooth, everywhere-differentiable, and bounded slightly above $0$ to avoid exploding gradients.
Though other works use the exponential function to map ``temperature logits'' $\in \mathbb{R}$ to $\mathbb{R}^{>0}$, we found that accidental exponential temperature growth could squash gradients.
When using exponentiation, a canonical instability arose when the network overstepped towards $\tau = 0$ due to momentum, then swung dramatically towards $\tau = \infty$. Here, it lost all gradients and could not recover.

\begin{figure}[h]
    \centering
    \includegraphics[width=.5\linewidth]{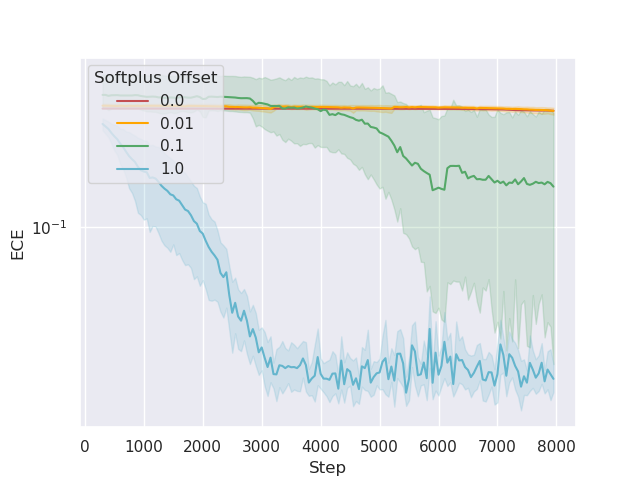}
    \caption{Learning curves for calibrating ResNet50 on Cifar100 with different softplus offsets. Having a slight offset in the function that transforms logits to temperatures or Gaussian scales greatly improves stability and convergence time. Without this offset, the optimization frequently diverged.}
    \label{fig:stability}
\end{figure}

To counteract this behavior we employ a shifted softplus function that is re normalized so $f(0) = 1$:

\begin{equation}
    \tau = f(x, s) = \frac{ln(1 + e^x) + s}{ln(2) + s}
    \label{eqn:shifted_softplus}
\end{equation}

where $s$ is a small constant that serves as a smooth lower bound to the temperature. This function decays to $s$ when $x < 0$, and increases linearly when $ x > 0$ , hence avoids the vanishing gradient problem. Figure \ref{fig:stability} demonstrates the importance of the offset in Equation \ref{eqn:shifted_softplus} with ResNet50 on Cifar100 calibration. We also found it important to normalize the features before processing with the layers dedicated to scales and temperatures. For shifted softmax offsets, we frequently employ $s=0.2$.

\section{Space and Time Characteristics of Likelihood Parameters}

\begin{table}[h]
\caption{
Space and time requirements of likelihood parameter types by the number of parameters $p$, 
training set size $n$,
feature dimension $d$, 
and the computation time for forward $f$, 
gradient $g$,
and storage $s$.
Space is measured for all points, while time is measured for one point.
}
\label{tab:constant-data-predicted}
\setlength{\tabcolsep}{2pt}
\begin{center}
\begin{tabular}[.35\textwidth]{lll}
Type & Space & Time \\ \hline
Global & $O(p)$ & $O(pg)$ \\
Data & $O(pn)$ & $O(p(f + g + s))$ \\
Predicted & $O(pd)$ & $O(pd(f + g))$ \\
\end{tabular}
\end{center}
\setlength{\tabcolsep}{2pt}
\end{table}

\section{Datasets}
\label{sec:datasets}
The regression datasets used are sourced from the UCI Machine Learning repository \citep{dua2017uci}, \citep{soil} (soil), and \citep{fried} (fried).
Inputs and targets are scaled to unit norm and variance prior to fitting for all regression experiments and missing values are imputed using scikit-learn's ``SimpleImputer'' \citep{pedregosa2011scikit}.
Large-scale image classification experiments leverage Tensorflow's Dataset APIs that include the SVHN, \citep{netzer2011reading}, ImageNet \citep{deng2009imagenet}, CIFAR-100, CIFAR-10 \citep{krizhevsky2009learning}, and CelebA \citep{liu2015faceattributes} datasets.
The datasets used in ridge and lasso experiments are 500 samples of 500 dimensional normal distributions mapped through linear functions with additive gaussian noise.
Linear transformations use $\text{Uniform}[1,2]$ weights and LASSO experiments use sparse transformations. For outlier detection experiments we leverage the Outlier Detection Data Sets (ODDS) benchmark \citep{rayana2016odds}.

\section{Models}
\label{sec:models}

\subsection{Constraint}
Many likelihood parameters have constrained domains, such as the normal variance $\sigma \in [0, \infty)$.
To evade the complexity of constrained optimization, we define unconstrained parameters $\phi_u$ and choose a transformation $t(\cdot)$ with inverse $t^{-1}(\cdot)$ to map to and from the constrained $\phi$.
For positivity, $\exp$/$\log$ parameterization is standard \citep{kendall2017uncertainties,kendall2018multi,saxena2019data}.
However, this parameterization can lead to instabilities and we use the softplus, $s^{+}(x) = \log(1 + \exp(x))$, instead. Shifting the softplus further improves stability (see Figure \ref{fig:stability}).
For the constrained interval $[a,b]$ we use affine transformations of the sigmoid $s(x) = \frac{1}{1 + \exp(-x)}$ \citep{barron2019loss}.

\subsection{Regularization}

Like model parameters, likelihood parameters can be regularized.
For unsupervised experiments we consider weight decay $\|\phi\|^2_2$, gradient clipping $\nabla_\phi / \|\nabla_\phi\|_2^2$, and learning rate scaling $\alpha_\phi = \alpha \cdot m$ for learning rate $\alpha$ and multiplier $m$.
We inherit the existing setting of weight decay for model parameters and clip at gradient norms at $1$.
We set the learning rate scaling multiplier to $m = 0.1$.





\subsection{Optimization}
The regressor parameters, $\eta$, are optimized by backpropagation through the regressed likelihood parameters $\phi = g_\eta(x)$.
The weights in $\eta$ are initialized by the standard Glorot \citep{glorot2010understanding} or He \citep{he2015delving} techniques with mean zero.
The biases in $\eta$ are initialized by the inverse parameter constraint function, $T^{-1}$, to the desired setting of $\phi$.
The default for variance and temperature is $1$, for equality with the usual squared error and softmax cross-entropy.

Regressor learning can be end-to-end or isolated.
In end-to-end learning, the gradient w.r.t. the likelihood is backpropagated to the regressor's input.
Whereas in isolated learning, the gradient is stopped at the input of the likelihood parameter model.
Isolated learning of predicted parameters is closer to learning global and data parameters, which are independent of the task model, and do not affect model parameters.

\subsection{Experimental Details}

Regression experiments utilize Keras' layers API with rectified linear unit (ReLU) activations and Glorot uniform initialization \citep{relu,glorot}. 
We use Keras implementations of DenseNet-121 \citep{huang2017densely} and ResNet-50 \citep{he2016deep} with default initializations. 
For image classifiers, we use Adam optimization with $lr=0.0001, \beta_1=.9, \beta_2=.99$ \citep{kingma2014adam} and train for 300 epoch with a batch size of 512.
Data parameter optimization uses Tensorflow's implementation of sparse RMSProp \citep{tieleman2012lecture}.
We train regression networks with Adam and $lr=0.001$ for 3000 steps without minibatching.
Deep regressors have a single hidden layer with 10 neurons, and recalibrated regressors have 2 hidden layers.
We constrain Normal variances and softmax temperatures using the affine softplus $s^{+}_{.01}$ and $s^{+}_{.2}$ respectively. 
Adaptive regularizer scales use $\exp$ parametization. 
We run experiments on Ubuntu 16.04 Azure Standard NV24 virtual machines (24 CPUs, 224 Gb memory, and 4$\times$ M60 GPUs) with Tensorflow 1.15 \citep{tensorflow}.

In our VAE experiments, our likelihood parameter model is a $1 \times 1$ convolution on last hidden layer of the decoder, which has the same resolution as the output.
The low and high dimensional losses use the same convolutional regressor, but the 1 dimensional case averages over pixels.
In the high dimensional case, the output has three channels (for RGB), with six channels total for shape and scale regression.
We use the same non-linearities to constrain the shape and scale outputs to reasonable ranges as in \citep{barron2019loss}: an affine sigmoid to keep the shape $\alpha \in [0, 3]$ and the softplus to keep scale $c \in [10^-8, \infty)$. Table \ref{tab:vae} gives the results of evaluating each method by MSE on the validation set, while training each method with their respective loss parameters.

For implementations of various outlier detection baselines we leverage the pyOD package \citep{zhao2019pyod} which provides all baselines with sensible default values. We use a Scikit-Learn's standard scaler to pre-process the data. Our approach leverages layers from the PyTorch API \citep{NEURIPS2019_9015}, and use a learning rate of $.0005$ for $4000$ steps with 20\% dropout before the code space.

\end{document}